\documentclass{article}
\usepackage{spconf,amsmath,graphicx, multirow}
\usepackage[numbers,sort&compress,comma]{natbib}

\usepackage{booktabs}
\usepackage{color}
\usepackage{bm}
\usepackage{balance}

\def \eg {\emph{e.g.}}
\def \ie {\emph{i.e.}}
\title{Attention-based dual-stream Vision Transformer for radar gait recognition}
\name{Shiliang Chen$^{\star}$ 
\qquad 
Wentao He$^{\star}$ 
\qquad
Jianfeng Ren$^{\star}$ 
\qquad
Xudong Jiang$^{\dagger}$ 
\thanks{This work was supported in part by the National Natural Science Foundation of China under Grant 72071116, and in part by the Ningbo Municipal Bureau Science and Technology under Grants 2019B10026. }
}

\address{$^{\star}$ The School of Computer Science, University of Nottingham Ningbo China \\
$^{\dagger}$ School of Electrical \& Electronic Engineering, Nanyang Technological University
}

\begin{document}
%
\maketitle
\begin{abstract}
Radar gait recognition is robust to light variations and less infringement on privacy. Previous studies often utilize either spectrograms or cadence velocity diagrams. While the former shows the time-frequency patterns, the latter encodes the repetitive frequency patterns. In this work, a dual-stream neural network with attention-based fusion is proposed to fully aggregate the discriminant information from these two representations. The both streams are designed based on the Vision Transformer, which well captures the gait characteristics embedded in these representations. The proposed method is validated on a large benchmark dataset for radar gait recognition, which shows that it significantly outperforms state-of-the-art solutions.

\end{abstract}
%

\section{Introduction}
\label{sec:intro}
Human gait recognition has become increasingly attractive in biometric applications such as public safety monitoring, health screening and human-computer interaction \cite{zhang2018latern}. Traditional gait recognition methods~\cite{liao2020model,zhang2016siamese} often require videos captured from a side view and heavily depend on lighting conditions. Taking pictures of people may cause privacy issues. In contrast, radar can capture the micro-Doppler signatures (mDS) of front-view gait features from a moving target~\cite{addabbo2021temporal}, revealing human dynamics and recognize human motions and gestures \cite{addabbo2021temporal,bai2019radar,le2018human}. More importantly, radar has less invasion of privacy and can work robustly in a variety of real-world situations such as dim conditions. In this paper, the problem of front-view human identification using radar mDS is studied.

Many mDS representations have been developed, \eg, spectrograms~\cite{ren2017regularized,ren2021three}, cadence velocity diagrams (CVD)~\cite{bjorklund2012evaluation} and cepstrograms~\cite{harmanny2014radar}. The spectrogram is a time-varying representation of mDS in the time-frequency domain \cite{addabbo2021temporal}, which has been applied to human activity recognition~\cite{le2018human,kim2015human}. The time-varying gait information such as the swinging speed of arms and legs can be well encoded in the spectrogram. However, the gait spectrograms of different people have very little differences, while the same person may have different gait characteristics, which makes the human identification using radar gait features particularly challenging. 
The CVD is another mDS representation by taking Fast Fourier Transform of the spectrogram along the time axis~\cite{bjorklund2012evaluation}. The CVD is less studied for gait recognition~\cite{seifert2017new}. It provides a useful measurement of repetition patterns for different velocities of body parts, which is a supplement to the spectrogram. 


Traditional classifiers such as Na\"{i}ve Bayes~\cite{fioranelli2016centroid}, support vector machines~\cite{begg2005support} and $k$-nearest-neighbour classifiers~\cite{van2017classification} have been used to classify mDS, while deep convolutional neural networks (DCNNs) often produce better performance. DCNNs have been used on spectrograms for radar gesture recognition and action classification \cite{addabbo2021temporal,le2018human,cao2018radar}, while other representations such as CVD are not fully explored but could provide complementary information to spectrogram. It is hence advantageous to use both spectrogram and CVD for classification. In addition, those DCNNs directly adopted from image recognition tasks often ignore the unique physical nature of the radar signal compared to optical images.

In this work, an Attention-based Dual-Stream Vision Transformer (ADS-ViT) is proposed to recognize people through radar gaits. As spectrograms often focus on the short-time-varying nature of mDS only, the CVD is introduced to capture the information on how often different frequencies repeat across a long duration. The proposed dual-stream network extracts features from these two representations simultaneously. Both streams are designed based on the Vision Transformer (ViT) to deeply exploit the gait characteristics embedded in patches of spectrogram and CVD. Furthermore, an attention-based fusion network is designed to optimally combine features from two streams. The proposed framework is validated on a large dataset for radar gait recognition, which shows that it significantly outperforms state-of-the-art models.

Our contributions are two-fold: 1) The proposed ADS-ViT effectively extracts and fuses features from spectrogram and CVD for radar gait recognition. 2) Both streams of ADS-ViT utilize the patch-processing ability of ViT to effectively capture the gait characteristics embedded in patches corresponding to frequency bands of spectrogram and CVD.


\section{Proposed attention-based dual-stream vision transformer }
\label{sec:methodology}
 \begin{figure*}[pt]
	\centering
	\includegraphics[width=1\textwidth]{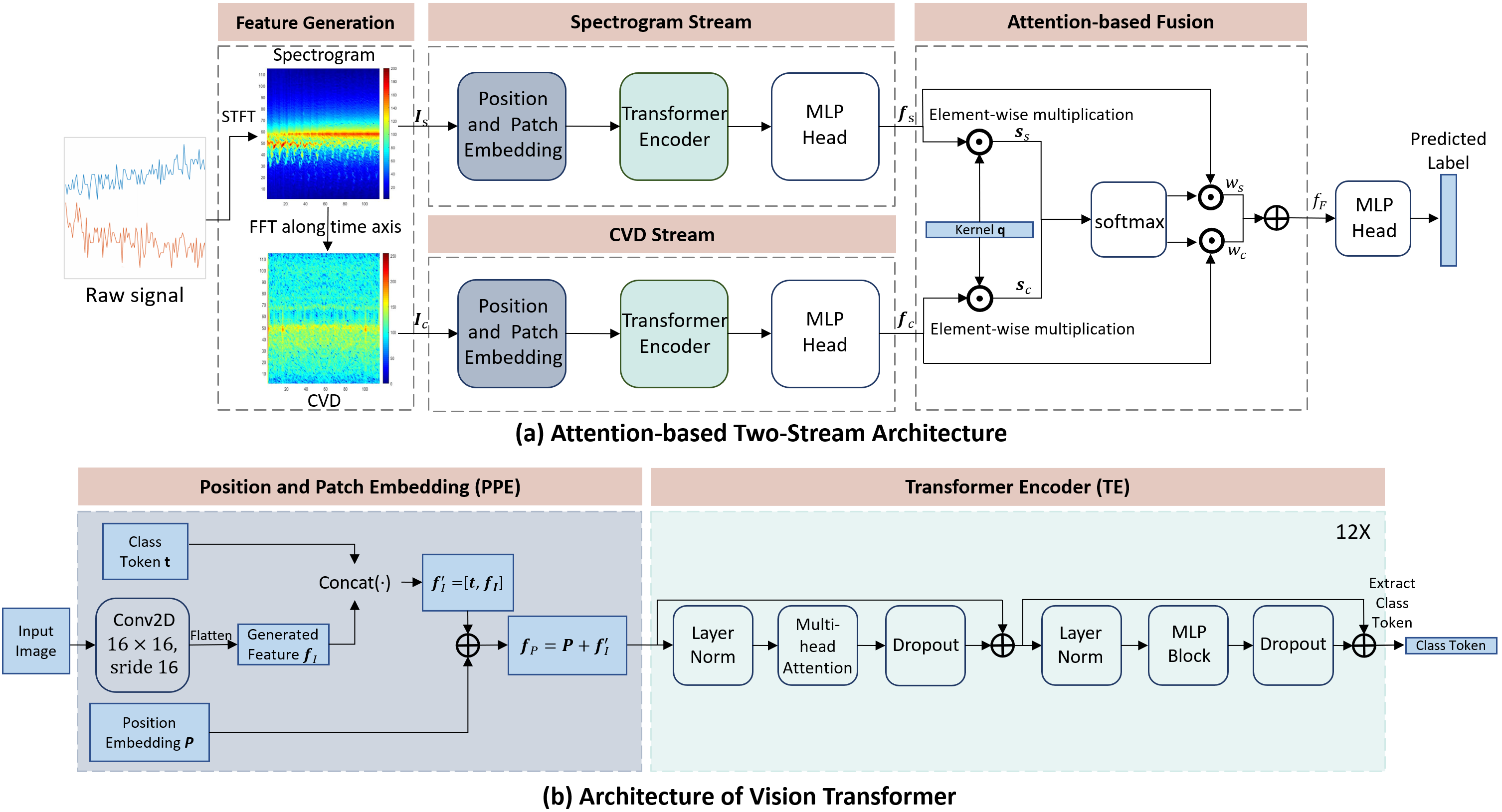}
	\caption{Overview of the proposed method shown in (a) and pipeline of ViT shown in (b).}
	\label{fig:block}
\end{figure*}


\subsection{Overview of Proposed Method}
The proposed ADS-ViT for radar gait recognition is shown in Fig. 1(a). To better exploit the discriminant information embedded in the radar signal, two initial feature representations, spectrogram~\cite{ren2017regularized,ren2021three} and CVD~\cite{bjorklund2012evaluation}, are jointly utilized in the proposed framework. Two subnetworks, spectrogram stream and CVD stream, are used to extract features from each of these two representations, respectively. For spectrograms and CVDs, different image patches correspond to different frequency bands. To robustly extract low-level features from these frequency bands, a network based on the Vision Transformer (ViT)~\cite{dosovitskiy2020image} is designed, where the local features are encoded through the Position and Patch Embedding mechanism of ViT. ViT could also capture the high-level semantic information across different frequency bands by utilizing the self-attention mechanism in the Transformer Encoder of ViT. An attention-based fusion mechanism is proposed to find the most discriminant features, and fuse the features of two streams into one feature map. Finally, a fully-connected layer is used to predict the person's identity given the radar signal.

More specifically, the proposed ADS-ViT network can be formulated as a quadruplet $\mathcal{Q} = (\mathcal{V}_s, \mathcal{V}_c, \mathcal{F}, \mathcal{C})$, where $\mathcal{V}_s$ and $\mathcal{V}_c$ are the ViT networks for spectrogram and CVD respectively, as shown in Fig.~\ref{fig:block}(b), $\mathcal{F}$ is an attention-based fusion network and $\mathcal{C}$ is the classifier at the end. Denote the generated spectrogram and CVD as $\bm{I}_s $ 
and $\bm{I}_c $, 
respectively. The features $\bm{f}_{s}$ from the spectrogram stream and $\bm{f}_{c}$ from the CVD stream are generated as:
\begin{equation}
	\bm{f}_{s} = \mathcal{V}_s(\bm{I}_{s}),
\end{equation}
\begin{equation}
	\bm{f}_{c} = \mathcal{V}_c(\bm{I}_{c}),
\end{equation}
These two sets of features are fused by the attention-based fusion network $\mathcal{F}$ to generate the features $\bm{f}_{F}$ 
as:
\begin{equation}
	\bm{f}_{F} = \mathcal{F}(\bm{f}_{s}, \bm{f}_{c}).
\end{equation}
Finally, the label $\hat{l}$ is predicted by the classifier $\mathcal{C}$ as:
\begin{equation}
\hat{l} = \mathcal{C}(\bm{f}_{F}).
\end{equation}

\subsection{Time-frequency Representations}
\label{sssec:complementary}
Spectrograms~\cite{ren2017regularized,ren2021three} and CVDs~\cite{bjorklund2012evaluation} are jointly utilized for robust radar gait recognition in this paper.

\noindent \textbf{Spectrogram:} Tahmoush and Silvious~\cite{ seifert2017new} modeled Doppler of each body part of a walking human as sinusoidal modulation in the spectrogram. This model reveals the velocities of each body part and determines the necessary human gait characteristics. The extracted features are the mean Doppler velocity and the size of torsos in the spectrogram. The mDS of a person is found relatively consistent and different walking people exhibit discriminative characteristics in the spectrogram~\cite{seifert2017new}. The radar spectrogram can well manifest the time-varying characteristics of a person's gait. A sample spectrogram is shown in Fig.~\ref{fig:block}(a).



\noindent \textbf{Cadence velocity diagram:} After deriving the spectrogram, the CVD~\cite{bjorklund2012evaluation} is obtained by taking the Fourier transform of the spectrogram along the time axis. The derived CVD, as shown in Fig.~\ref{fig:block}(a), is a matrix with rows representing Doppler frequencies and columns representing cadence frequencies, which measures how frequently different frequencies appear in the signal over the observation duration. It encodes useful information of the repetition of velocities, which would be the key feature to identify a walking person. 

\subsection{Vision Transformer for Spectrogram and CVD}
\label{sssec:ADS-ViT}
Two network branches based on the Vision Transformer are designed to extract features from spectrograms and CVDs, respectively. The ViT network consists of two parts: Position and Patch Embedding (PPE) and Transformer Encoder (TE), as shown in Fig. 1(b). The original size of spectrogram/CVD is $115\times115$. To fit the pretrained ViT network, the input spectrogram/CVD is converted to a 3-channel image $\bm{I} \in \mathcal{R}^{224\times224\times3}$, by replicating the resized grayscale image for three channels. In the PPE stage, $\bm{I}$ is convoluted into the feature map of size $14 \times 14 \times 768$  via a $16\times16$ kernel with a stride of 16, and flattened to the features $\bm{f}_I\in\mathcal{R}^{196\times768}$. 
Similarly as in~\cite{dosovitskiy2020image}, the learnable classification token $\bm{t}\in\mathcal{R}^{1\times768}$ is prepended to $\bm{f}_I$ as:
\begin{equation}
\bm{f}'_I = [\bm{t}, \bm{f}_I].
\end{equation}
The position information is embedded by adding a learnable position matrix $\bm{P}\in \mathcal{R}^{197 \times 768}$ for retaining positional knowledge of each patch, similarly as in~\cite{dosovitskiy2020image}, which results in the feature map $\bm{f}_P \in \mathcal{R}^{197 \times 768}$ with position embedding to the Transformer Encoder as:
\begin{equation}
\label{eq:addition}
    \bm{f}_P = \bm{P} + \bm{f}'_I.
\end{equation}
The TE module consists of 12 repeating blocks, where each contains a multi-head self-attention layer. This module could learn the global relations between patches. The learned knowledge is represented in the classification tokens for all patches and used as the output of the ViT network.

Unlike real-world images in which an object could be positioned anywhere in an image but interpreted the same, patches of spectrogram or CVD generated from the radar signal have the unique physical meanings when positioned at different locations in the image. Each patch in spectrogram or CVD contains the information in different frequency bands at different time instances (or different cadence velocities for CVD). Even patches similar in appearance could be physically different based on their frequency bands and temporal positions. 

The commonly-used convolutional neural networks such as AlexNet~\cite{cao2018radar}, VGG~\cite{addabbo2021temporal} and
ResNet~\cite{addabbo2021temporal} often use the same convolutional kernel across the whole image, while the proposed ADS-ViT treats patches differently, \ie, it splits the spectrogram/CVD into patches and embeds their positions so that the information embedded in different frequency bands could be properly aligned and effectively extracted. This partially justifies the superior performance of the proposed ADS-ViT over other models. The multi-head self-attention mechanism in the Transformer Encoder could further extract the reserved discriminant information from each patch and process this information globally to encode the global relations among patches.

\subsection{Attention-based Fusion}
\label{sssec:attention}
Inspired by Chen \emph{et al.} \cite{chen2019attention}, an attention-based fusion architecture shown in Fig.~\ref{fig:block}(a) is developed to fuse the complementary features extracted from both spectrogram and CVD. The target of the attention-based feature-level fusion is to find a set of weights $\{\bm{w}_i \in \mathcal{R}^{1\times768}\}$ for the features $\{\bm{f}_i\in \mathcal{R}^{1\times768}\}$ to obtain an aggregated feature $\bm{f}_a\in \mathcal{R}^{1\times768}$:
\begin{equation}
\bm{f}_a=\sum_{i=1}^{n}\bm{w}_i \odot \bm{f}_i,
\label{eq:aggregated}
\end{equation}
where $\odot$ denotes element-wise multiplication, $n$ is the number of features and $n=2$ in this paper. In the attention-based fusion model, the kernel $\bm{q}\in \mathcal{R}^{1\times768}$ is required to be trained. The process begins with the element-wise multiplication between the feature vector $\bm{f}_i$ and the kernel $\bm{q}$:
\begin{equation}
\bm{s}_i = \bm{q} \odot \bm{f}_i,
\end{equation}
where $\bm{s}_i\in \mathcal{R}^{1\times768}$ are the confidence scores for feature representations. A softmax function is then applied to $\bm{s}_i$ to assure that the derived weights $\sum_i \bm{w}_i=\bm{1}\in\mathcal{R}^{1\times768}$:
\begin{equation}
\bm{w}_i =e^{\bm{s}_i} \oslash \sum_{j}e^{\bm{s}_j},
\end{equation}
where $\oslash$ denotes element-wise division. The fused feature vector is finally generated using Eq.~(\ref{eq:aggregated}). This attention-based fusion could well highlight the most discriminant features by training the kernel function $\bm{q}$.






\section{Experimental results}
\label{sec:pagestyle}


\subsection{Dataset}
\label{sssec:dataset}
There is no publicly available dataset for radar gait recognition, and hence a K-MC1 radar transceiver and an ST200 evaluation system launched by RFbeam Microwave GmbH are employed to collect a radar gait dataset by the authors. There are two data collection sessions at least two weeks apart. In each session, each volunteer walks 10 sequences. (Some only participate in the first session.) The dataset consists of 1670 walking sequences from 98 volunteers. For each volunteer, 50\% of sequences are randomly selected as the training set and the rest are used for testing. In each sequence, a volunteer walks away from the radar along a corridor about 40 meters long, turns around and walks back towards the radar, lasting about 30 seconds. The sampling rate of the original signal is 125k. After a decimation of 64, the sampling rate is about 1.95k. The spectrogram is built using 128 sample points with an overlapping ratio of 90\%. After removing the cluster and non-informative high-frequency components, the spectrum has 115 data points. To enrich the dataset, each sequence is cut into multiple frames of 115 data points, with a stride of 10. As a result, a total number of 45,768 frames of size $115\times115$ are generated, in which 22,894 are used for training and 22,874 for testing.


\subsection{Experimental Settings}
\label{sssec:modelVariants}

AlexNet has been used as the backbone of the Deep Convolutional Neural Network on spectrograms of the radar signal to identify people~\cite{cao2018radar}. In~\cite{addabbo2021temporal}, VGG16~\cite{simonyan2014very} and ResNet18~\cite{he2016deep} have been used for radar gait recognition. These three approaches are implemented and evaluated on our benchmark dataset. 
As the proposed ADS-ViT utilizes both spectrogram and CVD, AlexNet~\cite{cao2018radar}, VGG16~\cite{addabbo2021temporal} and ResNet18~\cite{addabbo2021temporal} 
are applied on the CVD for comparison as well. 



The size of frames is $115\times115$ initially and resized to $224\times224$ to fit the networks. Stochastic Gradient Descent is used as the optimizer of all models with a momentum of $0.9$ and a weight decay of $5\times10^{-5}$. The cross-entropy loss function is used. For learning parameters, a linear learning rate warmup and decay is used with a learning rate scheduler, and the initial learning rate is set to $0.01$ for ResNet, $0.001$ for AlexNet and $0.0001$ for VGG. The batch size is 128 for all models. The proposed ADS-ViT initialize ViT networks with pretrained weights for both spectrogram and CVD streams. All the models are trained for 500 epochs and the optimal performance on the test set is reported.

\subsection{Comparisons to State-of-the-art Approaches}
\label{sssec:comcnn}
\begin{table}[pt]
	\centering
	\caption{Comparisons to state-of-the-art AlexNet~\cite{cao2018radar}, VGG16~\cite{addabbo2021temporal} and ResNet18~\cite{addabbo2021temporal}. These models are implemented and evaluated on both spectrograms and CVDs on our benchmark dataset for comparison. }
	\vspace{1mm}
		\begin{tabular}{l|l}
			\hline
			Method  &  Accuracy \\ 		
			\hline
			DCNN-AlexNet on Spectrogram \cite{cao2018radar} & 71.56\% \\ 
			VGG16 on Spectrogram \cite{addabbo2021temporal}  & 69.24\%\\
			ResNet18 on Spectrogram \cite{addabbo2021temporal}   & 85.56\% \\
			\hline
			DCNN-AlexNet on CVD &   72.44\% \\ 
			VGG16 on CVD &   79.83\%  \\ 
			ResNet18 on CVD &   80.73\% \\ 
			\hline
			\textbf{Proposed ADS-ViT} & \textbf{91.02\%} \\ 
			\hline
		\end{tabular}
	\label{tab:results}
\end{table}

The comparisons to state-of-the-art models are summarized in Table~\ref{tab:results}. Most existing models are applied on spectrograms, \eg, AlexNet~\cite{cao2018radar}, VGG16~\cite{addabbo2021temporal} and ResNet18~\cite{addabbo2021temporal}. As the proposed method utilizes both spectrogram and CVD, these three CNN models are applied on the CVD as well.

Table~\ref{tab:results} shows that the proposed ADS-ViT significantly outperforms all compared methods. Compared to the second-best model, ResNet18 on spectrogram~\cite{addabbo2021temporal}, the proposed method improves the classification accuracy from 85.56\% to 91.02\%. The achieved significant performance gain is attributed to the proposed ADS-ViT in two folds. First, by dividing the spectrogram/CVD into patches, the proposed ViT network extracts the local features from different frequency bands using the Position and Patch Embedding mechanism and extracts the global features across different frequency bands using the self-attention mechanism embedded in the transformer encoder.

Second, while the spectrogram encodes the short-duration information using the short-time Fourier transform, the long-duration repeating frequency patterns are captured by the CVD. By utilizing both spectrogram and CVD, the short-duration repetitive patterns in the spectrogram and the long-duration repetitive frequency patterns in the CVD are captured by the two streams of the proposed ADS-ViT, respectively. The proposed attention-based fusion scheme well integrates the discriminant information extracted from spectrogram and CVD, which leads to the superior performance of the proposed ADS-ViT.

\section{conclusion}
\label{sec:conclusion}
The proposed Attention-based Dual-Stream ViT well solves the problem of radar gait recognition. The radar signal can be represented as a spectrogram or a cadence velocity diagram. The complementary nature of these two representations motivates us to utilize both representations for better classification performance. We propose to use the Vision Transformer to split the spectrogram and CVD into patches and embed their positions so that the information embedded in different frequency bands could be effectively extracted and properly aligned. This enables the model to exploit the deep physical knowledge of the radar signal. The proposed attention-based fusion scheme integrates the discriminant features from the two representations. The proposed model is compared with state-of-the-art models on our benchmark dataset. The experimental results demonstrate that the proposed model significantly outperforms all compared methods.



\vfill\pagebreak

\balance
\bibliographystyle{IEEEbib}
\bibliography{refs}
\end{document}